\documentclass[letterpaper,twocolumn,fleqn]{article} 
\usepackage{epsfig}
\usepackage{subfigure}
\usepackage{epstopdf}
\usepackage{color}
\usepackage{multirow}
\usepackage{hhline}
\usepackage{color}
\usepackage{ist}

\title{Texture Segmentation Based Video Compression \\ Using Convolutional Neural Networks}

\author{Chichen Fu$^{\star}$, Di Chen$^{\star}$, Edward Delp$^{\star}$, Zoe Liu$^{\dagger}$, Fengqing Zhu$^{\star}$\\ $^{\star}$Purdue University\\ $^{\dagger}$Google, Inc. (United States)}

\date{} 

\begin{document} 

\maketitle 

\thispagestyle{empty} 


\begin{abstract}
There has been a growing interest in using different approaches to improve the
coding efficiency of modern video codec in recent years as demand for web-based video consumption increases. 
In this paper, we propose a model-based approach that uses texture analysis/synthesis 
to reconstruct blocks in texture regions of a video to achieve potential coding gains using the 
AV1 codec developed by the Alliance for Open Media (AOM). 
The proposed method uses convolutional neural networks to extract texture regions in a frame, which 
are then reconstructed using a global motion model. Our preliminary results
show an increase in coding efficiency while maintaining satisfactory visual quality.
\end{abstract}

\section{Introduction}
\label{sec:intro}
With the increasing amount of videos being created and consumed, better video compression tools are needed to provide fast transmission and high visual quality. 
Modern video coding standards utilize spatial and temporal redundancy in the videos to achieve high coding efficiency and high visual quality with motion compensation techniques and 2-D orthogonal transforms. However, efficient exploitation of statistical dependencies measured by a mean squared error (MSE) does not always produce the best psychovisual result, and may require higher data rate to preserve detail information in the video. 

Recent advancement in GPU computing has enabled the analysis of large scale data using deep learning method. Deep learning techniques have shown promising performance in many applications such as object detection, natural language process, and synthetic images generation \cite{krizhevsky2012, long2015, schmidhuber2015, lecun1998}.
Several methods have been developed for video applications to improve coding efficiency using deep learning. In \cite{park2016}, sample adaptive offset (SAO) is replaced by a CNN-based in-loop filter (IFCNN) to improve the coding efficiency in HEVC. By learning the predicted residue between the quantized reconstructed frames obtained after de-blocking filter (DF) and the original frames, IFCNN is able to reconstruct video frames with higher quality without requiring any bit transmission during coding process.
Similar to \cite{park2016}, \cite{Dai2017} proposes a Variable-filter-size Residue-learning CNN (VRCNN) to improving coding efficiency by replacing DF and SAO in HEVC. VRCNN is based on the concept of ARCNN \cite{Dong2015} which is originally designed for JPEG applications. 
Instead of only using spatial information to train a CNN to reduce the coding artifacts in HEVC, \cite{Jia2017} proposed a spatial temporal residue network (STResNet) as an additional in-loop filter after SAO. A rate-distortion optimization strategy is used to control the on/off switch of the proposed in-loop filter. 
There are also some works that have been done in the decoder of HEVC to improve the coding efficiency. In \cite{Wang2017}, a deep CNN-based auto decoder (DCAD) is implemented in the decoder of HEVC to improve the video quality of decoded video.  DCAD is trained to learn the predict residue between decoded video frames and original video frames.  By adding the predicted residual generated from DCAD to the compressed video frames, this method enhances the compressed video frames to higher quality. 

In summary, the above methods improve the coding efficiency by enhancing the quality of reconstructed video frames. However, they require different trained models for video reconstruction at different quantization levels.

We are interested in developing deep learning approaches to only encode visually relevant information and use a different coding method for ``perceptually insignificant" regions in a frame, which can lead to substantial data rate reductions while maintaining visual quality. In particular, we have developed a model based approach that can be used to improve the coding efficiency by identifying texture areas in a video frame that contain detail irrelevant information, which the viewer does not perceive specific details and can be skipped or encoded at a much lower data rate.  
The task is then to divide a frame into ``perceptually insignificant" texture region and then use a texture model for the pixels in that region. 

In 1959, Schreiber and colleagues proposed a coding method that divides an image into textures and edges and used it in image coding \cite{schreiber1959}. This work was later extended by using the human visual system and statistical model to determine the texture region \cite{peterson1990,kunt1985,delp1979}. More recently, several groups have focused on adapting perceptual based approaches to the video coding framework \cite{ndjiki-nya2012}.   
In our previous work \cite{bosch2011}, we introduced a texture analyzer before encoding the input sequences to identify detail irrelevant regions in the frame which are classified into different texture classes. At the encoder, no inter-frame prediction is performed for these regions. Instead, displacement of the entire texture region is modeled by just one set of motion parameters. Therefore, only the model parameters are transmitted to the decoder for reconstructing the texture regions using a texture synthesizer. Non-texture regions in the frame are coded conventionally.
Since this method uses feature extraction based on texture segmentation technique, a proper set of parameters are required to achieve accurate texture segmentation for different videos. 
Deep learning methods usually do not require such parameter tuning for inference. As a result, deep learning techniques can be developed to perform texture segmentation and classification for the proposed model-based video coding. A Fisher vector convolution neural networks (FV-CNN) that can produce segmentation labels for different texture classes was proposed in \cite{Cimpoi2015}. One of the advantage of FV-CNN is that the image input size is flexible and is not limited by the network architecture. Instead of doing pixel-wise classification on texture regions, a texture classification CNN network was described in \cite{Andrearczyk2016}. To reduce computational expenses, \cite{Andrearczyk2016} uses a small classification network to classify image patches with size of 227 $\times$  227. A smaller network is needed to classify smaller image patches in our case.
In this paper, we propose a block-based texture segmentation method to extract texture region in a video frame using convolutional neural networks. The block-based segmentation network classifies each 16 $\times$ 16  block in a frame as texture or non-texture.
The identified texture region is then synthesized using the temporal correlations among the frames. Our method was implemented using the AOM/AV1 codec. Preliminary results show significant bitrate savings while maintaining a satisfactory visual quality.   

\section{Method}
\begin{figure}[htb]
  \includegraphics[width=1\columnwidth]{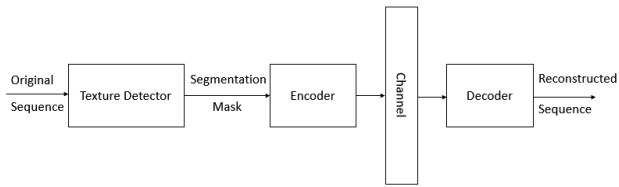}
  \caption{Block Diagram of Proposed Method}
  \label{fig:block_dia}
\end{figure}

Figure \ref{fig:block_dia} shows a block diagram of the proposed system. The original frames are first analyzed by texture detector to generate texture segmentation masks. Then, AOM/AV1 global motion tool is used to warping the identified texture region in a reference frame to synthesize the identified texture region in the current frame. The texture synthesis for each frame uses provided segmentation masks without sending residues for identified texture region. 


\subsection{Texture Analysis}
To provide texture information for the AOM/AV1 encoder, we use a block-based deep learning texture detector to analyze video frames and produce segmentation masks.
Our deep learning detector obtains texture segmentation masks by classifying each 16 $\times$ 16 block in a frame. 

\subsubsection{CNN Architecture}
\begin{figure}[htb]
	\includegraphics[width=1\columnwidth]{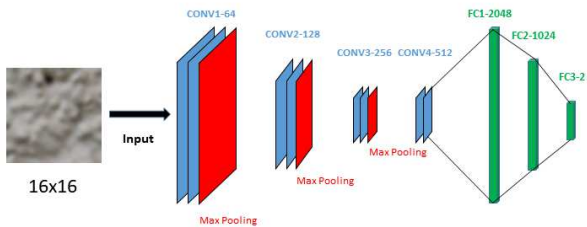}
	\caption{CNN Architecture for Texture Block Classification}
	\label{fig:archit}
\end{figure}

The CNN architecture of our method is shown in Figure \ref{fig:archit}.
Our CNN network is inspired by the VGG network architecture \cite{Simonyan2014}. The input of our architecture is a 16 $\times$ 16 color image block. The architecture consists of convolutional layers followed by a batch normalization rectified linear unit (ReLU) and a max pooling operation. Three fully connected layers with dropout operations and a softmax layer produces class probabilities. The output of our network is the probability that a 16 $\times$ 16 block is labeled as texture or non-texture, which indicates reliably of the texture/non-texture block label produced by the network. The kernel size of the convolutional layer is 3 $\times$ 3 and is padded by 1. The max pooling layer down samples the image by 2 and doubles the number of feature maps. 

\begin{figure}[htb]
\includegraphics[width=1\columnwidth]{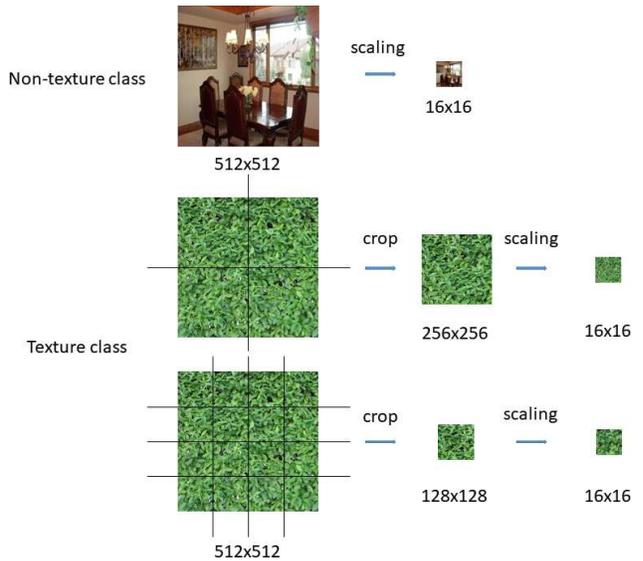}
\caption{CNN Training Data Preparation Procedure}
\label{fig:train}
\end{figure}

\subsubsection{Training}
The Salzburg Texture Image Database (STex) and the Places365 \cite{place365} were used for training the CNN. Images in the STex dataset are "pure" texture images. Images in the Place365 dataset are of general scenes. The texture class samples were created by cropping 512 $\times$ 512 STex images into 256 $\times$ 256 and 128 $\times$ 128 image patches and resizing them to 16 $\times$ 16 patches.  ThePlace365 images were resized into 16 $\times$ 16 image patches. Since the texture content in the Places365 images were lost during resizing operation, we can use resized Places365 images as non-texture class samples. In total, 1740 texture class images and 36148 non-texture class images were generated. The CNN training data preparation procedure is shown in Figure 2.

The proposed CNN architecture was implemented in Torch \cite{torch}. Mini-batch gradient descent is used with a fixed learning rate of 0.01, a momentum of 0.9 and a weight decay of 0.0005. The batch size of 512 image patches was trained in each iteration. For each epoch, 74 iterations were performed to cover the entire training set.  The training set was shuffled before each epoch. Since our training set is unbalanced for the texture class images and non-texture class images, the class weight of each class was set proportion to the inverse of class frequency. The error of training was calculated using cross entropy criterion and was converted to probability score using softmax regression. A total of 100 epochs were trained using one NVIDIA GTX Titan GPU.

\subsubsection{Inference}
After training the CNN, texture segmentation is performed using test video frames. Each frame is divided into 16 $\times$ 16 adjacent non-overlapping blocks. Each block in the video frames is classified as either texture or non-texture block. The segmentation mask for each frame is formed by grouping the classified blocks in the frame.

\subsection{Texture-Based Video Coding}
We use AOM/AV1 codec \cite{av1} to implement and test the proposed method.  
In our implementation, a video sequence is first divided into group of frames (GF group). 
In each GF group, texture synthesis method is disabled for the first frame and enabled for the rest of the frames. 
When a frame contains identified synthesizable texture regions, the texture region is synthesized by warping the corresponding region from the reference frames. 
This is implemented by using the global motion mode \cite{globalmotion} in AOM/AV1.
The texture analyzer is integrated into the encoder as follows and is illustrated in Figure \ref{fig:flow}:

\begin{figure}[htb]
  \includegraphics[width=1\columnwidth]{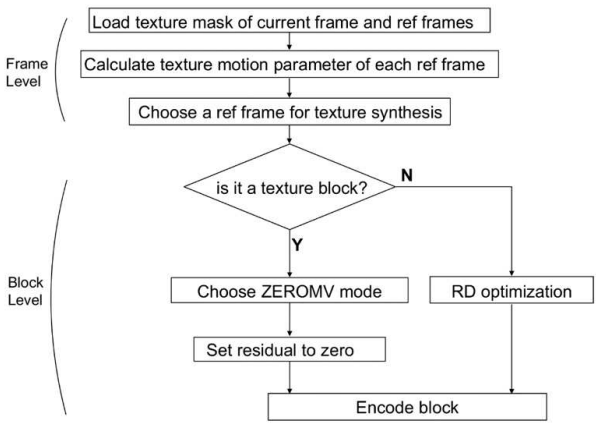}
  \caption{Encoding Operation Flow}
  \label{fig:flow}
\end{figure}

\begin{enumerate}
	\item Load and store texture mask for the current frame and all its reference frames;
	\item Estimate texture motion parameters for all the reference frames and replace at least one set of the codec's global motion parameters with the texture motion parameters;
	\item Check if a block is a texture block;
	\item Stop further block splitting for texture blocks;
	\item Choose global motion mode for texture blocks;
	\item Set the residual of texture block to be zero.
\end{enumerate}

\subsubsection{Texture Motion Parameters}
In order to get more accurate motion estimation of texture region, instead of using global motion parameters from the codec which are estimated based on the reference frame and the entire current frame, we obtained the texture motion parameters for each frame based on the reference frame and the texture region in the current frame. 
This indeed reduces the artifacts on the block edges of the texture blocks and non-texture blocks in the reconstructed video. 
The texture motion parameters are sent to the decoder in the uncompressed header of inter predicted frames. 

\subsubsection{Texture Block Decision} 
If a block is entirely inside the texture region of the current frame and the warped area in the reference frame is also inside the texture region of the reference frame, then this block is considered a texture block. 
If the block is a texture block, we do not further split it into smaller sub-blocks. If the block contains no texture region, RD optimization is performed for block partitioning and mode decision. 

\subsubsection{Texture Synthesis}
The global motion tool of AOM/AV1 codec was used to synthesize texture blocks by warping the reference frame. 
On the encoder side, the texture blocks select global motion mode and only use the reference frames that have texture motion parameters. 
To avoid artifacts on the block edges of the texture blocks, all texture blocks within one frame use the same reference frame. 
The residual of the texture blocks is set to zero.

\section{Experimental Result}
To evaluate our method, four pairs of data rates and PSNRs are calculated for each test sequences using the baseline codec. The baseline codec and our proposed method both use a single-layer coding structure. There are 4-16 frames in each GF group. Our proposed method enables texture mode for all the frames except the GOLDEN and ALTREF frames in AV1, which means that there is at least one frame in each GF group that does not use texture mode. 
Bjontegaard metric \cite{bdrate} is used to evaluate the performance of our method. Bjontegaard metric are BD-RATE and BD-PSNR which measure average data rate change and PSNRs change respectively.  The data rate is computed by dividing the WebM file size by the number of frames. The WebM file is the output of the encoder. PSNR is calculated based only on non-texture regions of the frame for the decoded video. We obtain four different pairs of data rates and PSNRs for each test sequence. The BD-RATE and BD-PSNR were calculated using these four pairs of data rate and PSNR points. 

\begin{figure*}[htb]
  \includegraphics[width=1.9\columnwidth]{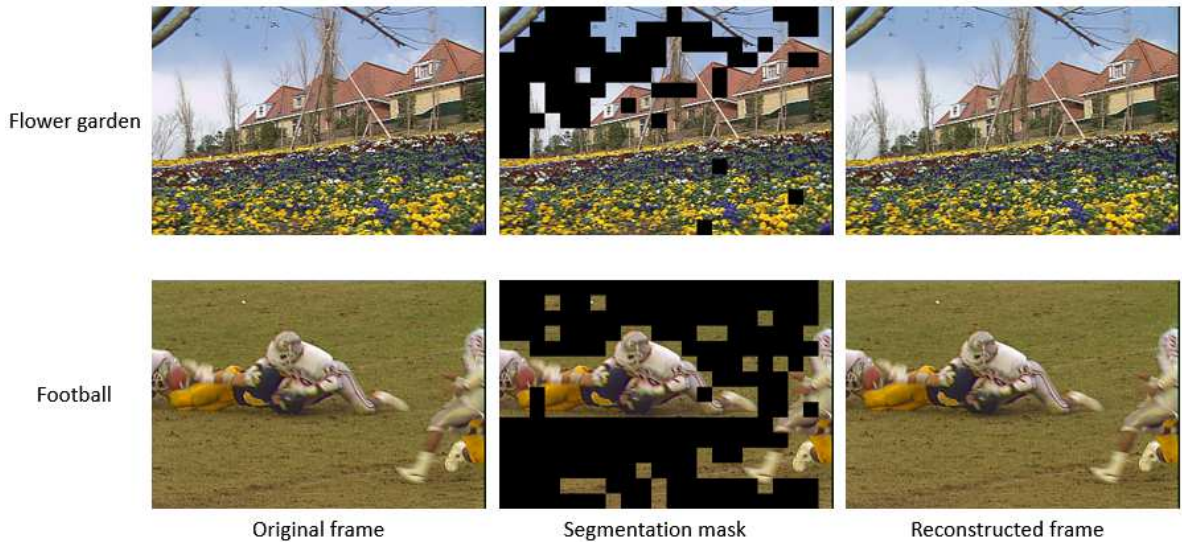}
  \caption{Sample Reconstructed Video Frames}
  \label{fig:recon}
\end{figure*}

\begin{figure*}[htb]
  \includegraphics[width=1.9\columnwidth]{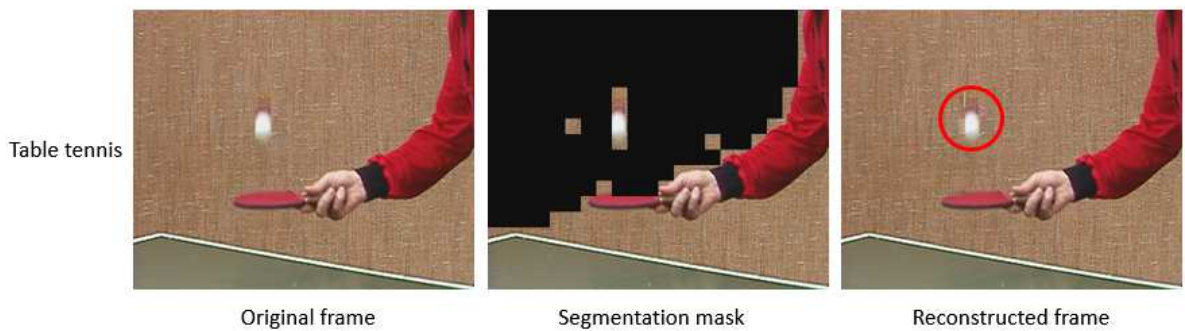}
  \caption{Sample Reconstructed Video Frame with Motion Artifacts Between the Edge of Texture Regions and Non-Texture Regions}
  \label{fig:artif}
\end{figure*}

\begin{table}[htb]
\centering
\caption{Table 1. BD-RATE and BD-PSNR}
\vspace{0.2in}
\renewcommand{\tabcolsep}{3pt}
{\begin{tabular}{|c|c|c|}
\hline

Sequence & BD-RATE (\%) & BD-PSNR (dB)\\
\hline
Flower garden & -16.49 & 1.25\\
\hline
Football & -5.68 & 0.42\\
\hline

\end{tabular}
}

\label{tab:BD}
\end{table}

\begin{table*}[htb]
\centering
\caption{Table 2. Data Rate Savings Obtained for Flower Garden Sequence}
\vspace{0.2in}
\renewcommand{\tabcolsep}{3pt}
{\begin{tabular}{|c|c|c|c|}
\hline

Quantization Level & Data Rate - AOM/AV1 (bits/frames) & Data Rate - Our Method (bits/frames) & Data Rate Savings (\%) \\
\hline
16 & 115330 & 136080 & 15.24\\
\hline
24 & 81558 & 94695 & 13.87\\
\hline
28 & 63621 & 73768 &13.76\\
\hline
32 & 51326 & 59326& 13.48 \\
\hline

\end{tabular}
}

\label{tab:flower}
\end{table*}

\begin{table*}[htb]
\centering
\caption{Table 3. Data Rate Savings Obtained for Football Sequence}
\vspace{0.2in}
\renewcommand{\tabcolsep}{3pt}
{\begin{tabular}{|c|c|c|c|}
\hline

Quantization Level & Data Rate - AOM/AV1 (bits/frames) & Data Rate - Our Method (bits/frames) & Data Rate Savings (\%) \\
\hline
16 & 112720 & 122110 & 7.69\\
\hline
24 & 79621 & 83874 & 5.71\\
\hline
28 & 62905 & 65811 & 4.42\\
\hline
32 & 49600 & 51705& 4.71 \\
\hline

\end{tabular}
}

\label{tab:football}
\end{table*}

Figure \ref{fig:recon} shows the reconstructed sample video frames. The quantitative evaluation results are shown in Table \ref{tab:BD}, \ref{tab:flower}, \ref{tab:football}. The evaluation results show large data rate saving for our sample decoded videos while the reconstructed video frames showing no significant visual artifacts.  For sequence with fast motion, motion artifacts are observed between the edge of texture regions and non-texture regions illustrated in Figure \ref{fig:artif}. In our current implementation, texture synthesis is performed using the codec's built-in global motion function. Inaccurate motion parameters can lead to motion artifacts in some sequences. A potential approach to fix the artifacts is to use different motion models when estimating the motion parameters. By addressing what types of motion may cause such visual artifacts, the most suitable motion model can be selected to fit the scheme in different videos. The complexity of the motion models used for each identified texture regions can be adjusted depending on the texture content. In addition, we would like to categorize different types of motion present in the video and assess how well our texture based method performs for different types of motion. This can then be used as feedback to automatically determine which mode is best suited for encoding the frame/block using machine learning methods.  

\begin{figure}[htb]
  \includegraphics[width=0.9\columnwidth]{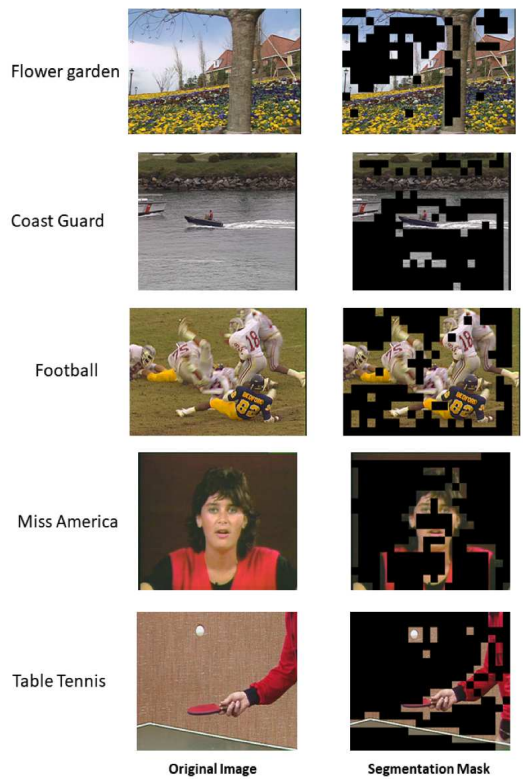}
  \caption{Sample Texture Segmenatation Results}
  \label{fig:texture}
\end{figure}

Figure \ref{fig:texture} shows the texture segmentation results. Our CNN detector is more likely to find texture blocks that only contain one single type of texture within a 16 $\times$ 16 block. If a block contains different types of texture content, it may not be detected as a texture block. 
Using more training samples that contains different types of textures could improve the performance of our CNN detector.  
Additional post-processing techniques such as connected components can be used to remove small texture regions.

\section{Conclusion}
In this paper, we proposed a texture segmentation based video coding method for AOM/AV1 codec. A deep learning based texture segmentation method was developed to detect texture regions in a frame that is ``perceptually insignificant." The proposed method is implemented in the AOM/AV1 codec by enabling the global motion mode to ensure temporal consistency. Preliminary results showed a large data rate saving without noticeable visual artifacts. 

{\small
\bibliographystyle{ieee}
\bibliography{egbib}

\begin{thebibliography}{10}
\providecommand{\url}[1]{#1}
\def\UrlFont{\rmfamily}
\providecommand{\newblock}{\relax}
\providecommand{\bibinfo}[2]{#2}
\providecommand\BIBentrySTDinterwordspacing{\spaceskip=0pt\relax}
\providecommand\BIBentryALTinterwordstretchfactor{4}
\providecommand\BIBentryALTinterwordspacing{\spaceskip=\fontdimen2\font plus
\BIBentryALTinterwordstretchfactor\fontdimen3\font minus
  \fontdimen4\font\relax}
\providecommand\BIBforeignlanguage[2]{{%
\expandafter\ifx\csname l@#1\endcsname\relax
\typeout{** WARNING: IEEEtran.bst: No hyphenation pattern has been}%
\typeout{** loaded for the language `#1'. Using the pattern for}%
\typeout{** the default language instead.}%
\else
\language=\csname l@#1\endcsname
\fi
#2}}

\bibitem{krizhevsky2012}
A.~Krizhevsky, I.~Sutskever, and G.~E. Hinton, ``{ImageNet} classification with
  deep convolutional neural networks,'' \emph{Proceedings of the Neural
  Information Processing Systems}, pp. 1097--1105, December 2012, {Lake Tahoe,
  NV}.

\bibitem{long2015}
J.~Long, E.~Shelhamer, and T.~Darrell, ``Fully convolutional networks for
  semantic segmentation,'' \emph{Proceedings of the IEEE Conference on Computer
  Vision and Pattern Recognition}, pp. 3431--3440, June 2015, {Boston, MA}.

\bibitem{schmidhuber2015}
J.~Schmidhuber, ``Deep learning in neural networks: An overview,'' \emph{Neural
  networks}, vol.~61, pp. 85--117, 2015.

\bibitem{lecun1998}
Y.~LeCun, L.~Bottou, Y.~Bengio, and P.~Haffner, ``Gradient-based learning
  applied to document recognition,'' \emph{Proceedings of the IEEE}, vol.~86,
  no.~11, pp. 2278--2324, November 1998.

\bibitem{park2016}
W.~Park and M.~Kim, ``Cnn-based in-loop filtering for coding efficiency
  improvement,'' \emph{Image, Video, and Multidimensional Signal Processing
  Workshop}, pp. 1--5, 2016.

\bibitem{Dai2017}
Y.~Dai, D.~Liu, and F.~Wu, ``A convolutional neural network approach for
  post-processing in hevc intra coding,'' \emph{International Conference on
  Multimedia Modeling}, pp. 28--39, 2017.

\bibitem{Dong2015}
C.~Dong, Y.~Deng, C.~C. Loy, and X.~Tang, ``Compression artifacts reduction by
  a deep convolutional network,'' \emph{Proceedings of the IEEE International
  Conference on Computer Vision}, pp. 576--584, 2015.

\bibitem{Jia2017}
C.~Jia, S.~Wang, X.~Zhang, S.~Wang, and S.~Ma, ``Spatial-temporal residue
  network based in-loop filter for video coding,'' \emph{arXiv preprint}, p.
  arXiv:1709.08462, 2017.

\bibitem{Wang2017}
T.~Wang, M.~Chen, and H.~Chao, ``A novel deep learning-based method of
  improving coding efficiency from the decoder-end for hevc,'' \emph{Data
  Compression Conference}, pp. 410--419, 2017.

\bibitem{schreiber1959}
W.~F. Schreiber, C.~F. Knapp, and N.~D. Kay, ``Synthetic highs, an experimental
  tv bandwidth reduction system,'' \emph{J. Soc. Motion Picture Televis. Eng.},
  vol.~68, pp. 525--537, 1959.

\bibitem{peterson1990}
H.~Peterson, ``Image segmentation using human visual system properties with
  applications in image compression,'' \emph{Ph.D. dissertation, Purdue Univ},
  May 1990, {West Lafayette, IN}.

\bibitem{kunt1985}
M.~Kunt, A.~Ikonomopoulos, and M.~Kocher, ``Second{-}generation image{-}coding
  techniques,'' \emph{Proc. IEEE}, vol.~73, no.~4, pp. 549--574, April 1985.

\bibitem{delp1979}
E.~J. Delp, R.~L. Kashyap, and O.~Mitchell, ``Image data compression using
  autoregressive time series models,'' \emph{Pattern Recognit.}, pp. 313--323,
  June 1979.

\bibitem{ndjiki-nya2012}
P.~Ndjiki-Nya, D.~Doshkov, H.~Kaprykowsky, F.~Zhang, D.~Bull, and T.~Wiegand,
  ``Perception-oriented video coding based on image analysis and completion: A
  review,'' \emph{Signal Processing: Image Communication}, vol.~27, no.~6, pp.
  579--594, July 2012.

\bibitem{bosch2011}
M.~Bosch, F.~Zhu, and E.~J. Delp, ``Segmentation{-}based video compression
  using texture and motion models,'' \emph{IEEE Journal of Selected Topics in
  Signal Processing}, vol.~5, no.~7, pp. 1366--1377, 2011.

\bibitem{Cimpoi2015}
M.~Cimpoi, S.~Maji, and A.~Vedaldi, ``Deep filter banks for texture recognition
  and segmentation,'' \emph{Proceedings of the IEEE Conference on Computer
  Vision and Pattern Recognition}, pp. 3828 -- 3836, 2015.

\bibitem{Andrearczyk2016}
V.~Andrearczyk and P.~F. Whelan, ``Using filter banks in convolutional neural
  networks for texture classification,'' \emph{Pattern Recognition Letters},
  pp. 63--69, 2016.

\bibitem{Simonyan2014}
K.~Simonyan and A.~Zisserman, ``Very deep convolutional networks for
  large-scale image recognition,'' \emph{arXiv preprint}, p. arXiv:1409.1556,
  2014.

\bibitem{place365}
B.~Zhou, A.~Khosla, A.~Lapedriza, A.~Torralba, and A.~Oliva, ``Places: An image
  database for deep scene understanding,'' \emph{arXiv preprint}, p.
  arXiv:1610.02055, 2016.

\bibitem{torch}
R.~Collobert, K.~Kavukcuoglu, and C.~Farabet, ``Torch7: A matlab-like
  environment for machine learning,'' \emph{Proceedings of the BigLearn
  workshop at the Neural Information Processing Systems}, pp. 1--6, December
  2011, {Granada, Spain}.

\bibitem{av1}
``Alliance for open media, press release online,''
  \url{http://aomedia.org/press-release}.

\bibitem{globalmotion}
S.~Parker, Y.~Chen, D.~Barker, P.~D. Rivaz, and D.~Mukherjee, ``Global and
  locally adaptive warped motion compensation in video compression,''
  \emph{Proc. IEEE Int. Conf. Image Process. (ICIP)}, September 2017.

\bibitem{bdrate}
G.~Bjøntegaard, ``Calculation of average psnr differences between rdcurves,''
  \emph{VCEGM33, 13th VCEG meeting}, March 2001, {Austin, Texas}.

\end{thebibliography}
}

\begin{biography}

Chichen Fu received his BS in Electrical Engineering from Purdue University (2014) 
and he is pursuing his PhD in Electrical and Computer Engineering from Purdue University. 
His research interest include machine learning, image processing and computer vision. 

Di Chen is a PhD candidate in Video and Image Processing Laboratory (VIPER) at Purdue University, West Lafayette. Her research focuses on video analysis and compression. Currently, she is working on texture segmentation based video compression using convolutional neural networks. 

Edward J. Delp was born in Cincinnati, Ohio. He is currently The Charles 
William Harrison Distinguished Professor of Electrical and Computer 
Engineering and Professor of Biomedical Engineering at Purdue 
University. His research interests include image and video processing, 
image analysis, computer vision, image and video compression, multimedia 
security, medical imaging, multimedia systems, communication and 
information theory. Dr. Delp is a Life Fellow of the IEEE, a Fellow of 
the SPIE, a Fellow of IS\&T, and a Fellow of the American Institute of 
Medical and Biological Engineering.

Zoe Liu is a software engineer with Google's WebM team and has been a key contributor to the open source video codec standard AOMedia Codec One (AV1). Zoe received her PhD from Purdue University and her ME and BE from Tsinghua University, Beijing. Either as a principal contributor or as a Technical Lead, Zoe had previously devoted her effort to the design and development of several video call products, including FaceTime, Tango Video Call, and Google Glass Video Call. Her main research interests include video compression, processing, and communications.

Fengqing Zhu is an Assistant Professor of Electrical and
Computer Engineering at Purdue University, West Lafayette, IN.
Dr. Zhu received her Ph.D. in Electrical and Computer Engineering
from Purdue University in 2011. Prior to joining Purdue
in 2015, she was a Staff Researcher at Huawei Technologies
(USA), where she received a Huawei Certification of Recognition
for Core Technology Contribution in 2012. Her research interests
include image processing and analysis, video compression,
computer vision and computational photography.

\end{biography}

\end{document}